# AUTOMATIC FEATURE HIGHLIGHTING IN NOISY RES DATA WITH CYCLEGAN

## A PREPRINT

**Nicholas Khami**
College of Natural Sciences, The University of Texas at Austin

**Omar Imtiaz***
College of Natural Sciences, The University of Texas at Austin

**Akif Abidi**
College of Natural Sciences, The University of Texas at Austin

**Akash Aedavelli**
College of Natural Sciences, The University of Texas at Austin

**Alan Goff**
College of Natural Sciences, The University of Texas at Austin

**Jesse R. Pisel**
Paul M. Rady School of Computer Science and Engineering, The University of Colorado at Boulder

**Michael J. Pyrcz**
Cockrell School of Engineering, Jackson School of Geosciences, The University of Texas at Austin

August 18, 2021

*Corresponding Author: oaimtiaz@utexas.edu*

## ABSTRACT

Radio echo sounding (RES) is a common technique used in subsurface glacial imaging, which provides insight into the underlying rock and ice without using destructive measurement techniques. However, systematic noise is often introduced into the data during collection, thereby complicating human interpretation of the results. Researchers most often use a combination of manual interpretation and filtering techniques to denoise data; however, these processes are both time intensive and may not always yield a desirable post-processed result. Fully Convolutional Networks (FCNs) have been proposed as an automated alternative to identify layer boundaries in radargrams. However, they are limited in that they require ground-truth training images, require high-quality manually processed training data, and struggle to interpolate data in noisy samples (Varshney et al. 2020).

Herein, the authors propose a GAN based model to interpolate layer boundaries through noise and highlight layers in two-dimensional glacial RES data. In real-world noisy images, filtering often results in loss of data such that interpolating layer boundaries is nearly impossible. Furthermore, traditional machine learning approaches are not suited to this task because of the lack of paired clean and noisy data, so we employ an unpaired image-to-image translation model. For this model, we create a synthetic dataset to represent the domain of "ideal" images with clear, highlighted layers and use an existing real-world RES dataset as our noisy domain.

We implement a CycleGAN trained on these two domains to highlight layers in noisy images that can interpolate effectively without significant loss of structure or fidelity. Though the current implementation is not a perfect solution, the model clearly highlights layers in noisy data and allows researchers to determine layer size and position without requiring mathematical filtering, manual processing, or ground-truth images for training. This is significant because clean images generated by our model enables subsurface researchers to determine glacial layer thickness more efficiently.

**1. Introduction**

Radio echo sounding (RES) data is useful to subsurface researchers in determining rock and ice layer thickness and position. Researchers find it useful in studying ice thickness, bed topography, and subglacial lakes. However, it often contains noise from instrumentation and environmental conditions. This interference is called strip noise and is best characterized as vertical, horizontal, and diagonal stripes that appear in RES radargram records (Figure 1). Strip noise makes it more difficult to determine the thickness of glacial layers as it obfuscates boundaries – such as the snow to ice transition zone. Tracking the properties of glaciers is important in climate monitoring research as their fluctuations serve as evidence of a change in climate. With the acceleration of climate change, climate monitoring is becoming increasingly relevant (Thompson 2010).

RES data is collected using ice penetrating radar systems such as ground penetrating radar systems, chirped airborne sounders, or electromagnetic based polarimetric methods. Like most physical measurement tools, RES data is susceptible to noise. In this paper, we will focus on 'strip noise'. Vertical and horizontal strip noise originates from internal instrument instability while point object cause diagonal strip noise. The removal of strip noise from RES data is the subject of many research papers (Huang et al. 2017). Current approaches use various filtering techniques (Chen et al. 2017; Qihua & Jing, 2012) or FCNs (Varshney et al. 2020). Filtering techniques are limited in that they only amplify strong signals, do not interpolate layer boundaries, and risk over-processing. Additionally, FCNs are limited in that they require significant manual interpretation of training data and cannot highlight layers. Figure 2 shows an example of filtered RES data that contains diagonal strip noise. In Figure 2 the glacial radargram is filtered in ImpDAR (Lilien et al. 2020) with a vertical and horizontal bandpass filter – the standard for ground penetrating radargrams (Hinterleitner 2009). The processed example shows that the filtering process removes the background coloring and forces researchers to rely on contrasting signal bands to manually identify layers. These boundaries contain multiples, obfuscating the precise location of the layer boundaries.

Because of these flaws in filtered data, we propose a model to produce an entirely different kind of output that removes intra-layer boundaries and highlights homogenous layers. This is beneficial to researchers as it makes the location of the layer boundaries apparent, so that average layer height can be more accurately and efficiently measured. To achieve these results, the model needs both real world and synthetic data to train.

Until now, no one has attempted to use a Generative Adversarial Network (GAN) to remove strip noise and isolate layers in RES data. This is likely because there are no paired datasets of noisy and human interpreted RES records. If the paired dataset consists of filtered data, deep-learning models would be inherently limited, only replicating the filtering process. Meaning that the deep-learning model would learn the underlying human assumptions that set the filtering parameters. Understanding this, we propose an unsupervised learning approach that employs a CycleGAN (Cycle Generative Adversarial Network) trained on unpaired RES records.

The GAN architecture is based on the idea of an adversarial loss that simulates competition between a generator network and a discriminator network. The adversarial loss is used to train the generator to create replicas of the training data. Specifically, CycleGAN is a GAN for image-to-image translation between two unpaired domains X and Y (Zhu et al. 2017). This translation process is illustrated in Figure 3. The model works by training two generator networks, G and F, where G translates images from domain X to domain Y, and vice versa for F. The outputs for both generators are analyzed by two discriminators DX and DY, where DY incentivizes G to better translate domain X images into domain Y images, and vice versa. The output from each generator is then fed into the other generator, forming a cycle of training. This technique is highly effective in scenarios where data is unpaired, as each generator learns to translate images to the alternate domain. Given the lack of paired noisy and interpreted RES records, a CycleGAN is an intuitive approach to translate noisy RES records into an interpreted form automatically.

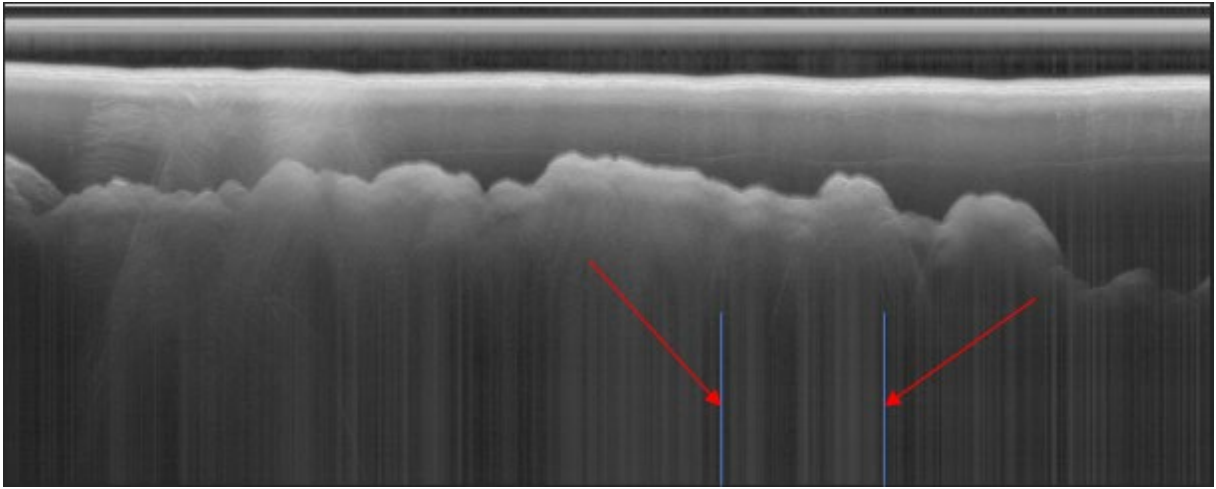

Figure 1: A visual representation of the IceBridge HiCARS 1 L1B Time-Tagged Echo Strength Profiles (Siegert, & Holt, 2017). The vertical lines in blue highlighted by the arrows are examples of strip noise that pollute RES data.

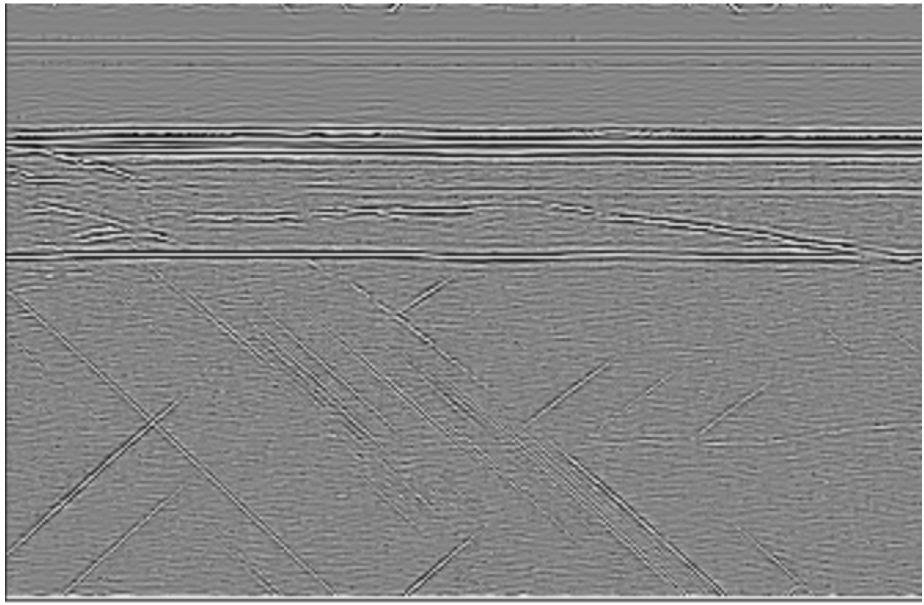

Figure 2: Filtered data from 1-DT Filtering. The filtering method helps exaggerate the different instances of diagonal strip noise in the original radargram image.

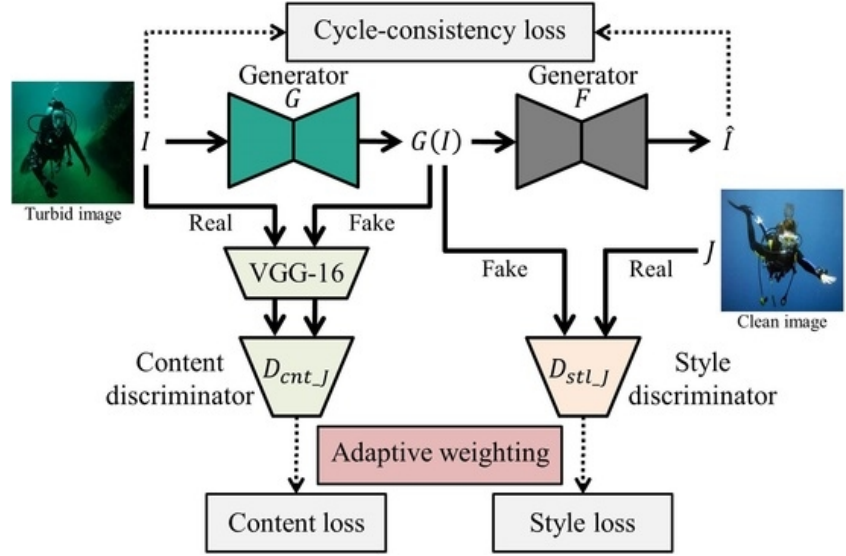

Figure 3: A sample run-through of a CycleGAN translating turbid underwater images (domain X) into clean underwater images (domain Y) (from Mao and Li 2020).

## 2. Methods

As stated above, CycleGAN requires two domains of data. For this study we use one real-world unfiltered dataset for the unprocessed domain, and one synthetic dataset for the interpreted domain. The real-world data comes from the National Snow & Ice Data Center (NSIDC); it contains Antarctic RES profiles from the HiCARS Version 1 instrument (Siegert, & Holt, 2017). The dataset is minimally preprocessed meaning it is the noisy domain of data. The dataset is high quality, open access, and is the subject of prior research on RES strip noise removal (Wang et al. 2019).

For visual quality control of data, we turn the RES records into radargram images. We first use the netCDF4 module (Whitaker 2021) to access the raw data arrays. Then, to transform the arrays into greyscale images, we scale the elements to values between 0 to 255. Following that transformation, we save the data as radargram images as shown in Figure 1.

Next, we create synthetic data, exemplified in figure 4, to define the clean domain. To create synthetic data, we use the wedge package within the Bruges library (Bianco et al. 2021). With this package, we randomize the different model parameters to increase the diversity of our generated data. This randomization helps to mitigate potential training errors due to structural differences between the clean and noisy data.

Through visual analysis, we validate that the type of data shown in Figure 4 is ideal for representing our target output as it contains layer boundaries with clear stratigraphy. Specifically, this data type is relatively clear compared to filtered radargrams like Figure 2 as it does not contain reflected signal bands and visually differentiates between layers.

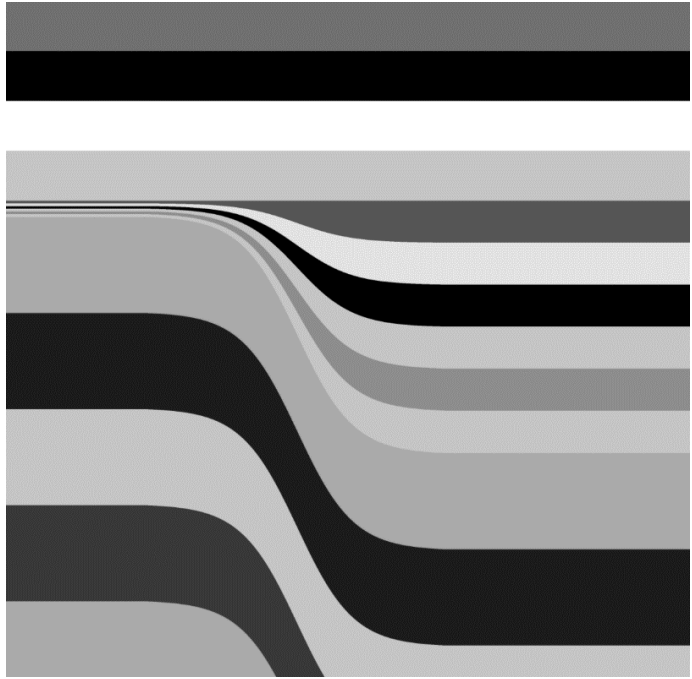

Figure 4: Clean Data from Bruges used as the clean data portion when training the CycleGAN, where the network will to convert real data to something similar.

To train the CycleGAN we use the synthetic data as domain X and the noisy data as domain Y. During training, the model uses the standard CycleGAN adversarial and cycle consistency losses (Zhu et al. 2017). To track model accuracy during training we use the Peak-Signal-To-Noise-Ratio (PSNR) and Mean Squared Error (MSE). The MSE is the cumulative squared error between two images while the PSNR — measured in decibels — is used as a quality measurement between two images. Using the initial and final images for a full translation cycle from domains Y to X and back to Y, we calculate the MSE and PSNR. PSNR is correlated with MSE, however more easily interpreted, therefore both values are included in this paper. These scores quantitatively measure the extent to which data is maintained in the translation process through a whole cycle (Sharma et al. 2019). After training we use a modified a loss function to validate our results.

The main issue we expected to have with the model was that it would erroneously add features to the data. However, we anticipated this would be manageable through tuning hyperparameters and manipulating the structure of the synthetic training data to resemble the real-world data more closely.

The model hyperparameters consist of learning rate, crop size, epoch count, batch size, instance normalization, and batch normalization. We manually optimize the hyperparameters varying the number of epochs varied from 100 to 500, crop size from 250px to 480px, learning rate from 0.001 to 0.0001, batch size from 1 to 3, instance normalization active or inactive and batch normalization active or inactive. This consists of approximately 20 model training runs.

## 3. Results

Optimal hyperparameters for the CycleGAN as follows: ResNet 9 blocks, a non-scaled random 400x100 pixel crop size, single channel preprocessing, batch normalization, 500 training epochs with a learning rate of 0.0001, batch size of two, and instance and batch normalization at each layer. Instance normalization reduces the variance in the values as the network trains while batch normalization serves as an alternative to increasing the image crop size. Meaning the model can train on more layers in domain Y crops during training while using less GPU memory. After finalizing the hyperparameters, we train the model for 500 epochs, saving a checkpoint every five. Theoretically, the model progressively eliminates more and more detail from the ground truth images and this overfitting is prevented by using checkpoints.

As glacial RES data is primarily used for measuring depth of subsurface layers, the result from epoch 500 was best – both heuristically and quantitatively based on our loss function – as it had the clearest, most interpretable results but still retained structural detail and effectively highlighted layers. Results after translating the noisy domain X to the clean domain Y are shown in Figures 5 through 8.

### 3.1 Model Performance

Figures 5 and 6 demonstrate the model's performance on RES data with minimal diagonal strip noise. Through detailed visual analysis, we can see that the CycleGAN is effective at parsing through the concentrated signal bands, identifying boundaries, and contrasting different layers. Yet, the model does not do well at maintaining the detail in the ice-bedrock/ice-water interface.

Figures 7 and 8 demonstrate the model's performance on RES data with a large amount of diagonal strip noise. As with Figures 5 and 6, the CycleGAN is effective at parsing through the concentrated signal bands, identifying boundaries, and contrasting different layers. Additionally, with this image type, the model still does a poor job of retaining detail in the ice-bedrock layer.

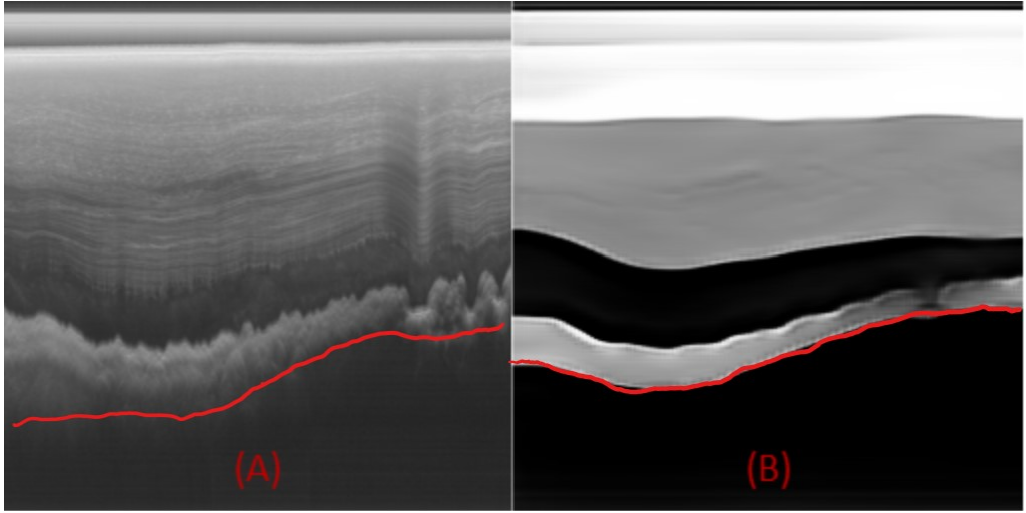

Figure 5: IR1HI1B_2009006_MCM_JKB1a_WLKX01a_005_HIGH. Image A is the original RES data, B is the original data transformed by the CycleGAN to imitate the Bruges data. The red line shows where the ice-bedrock/ice-water interface is according to the radargram and CycleGAN respectively.

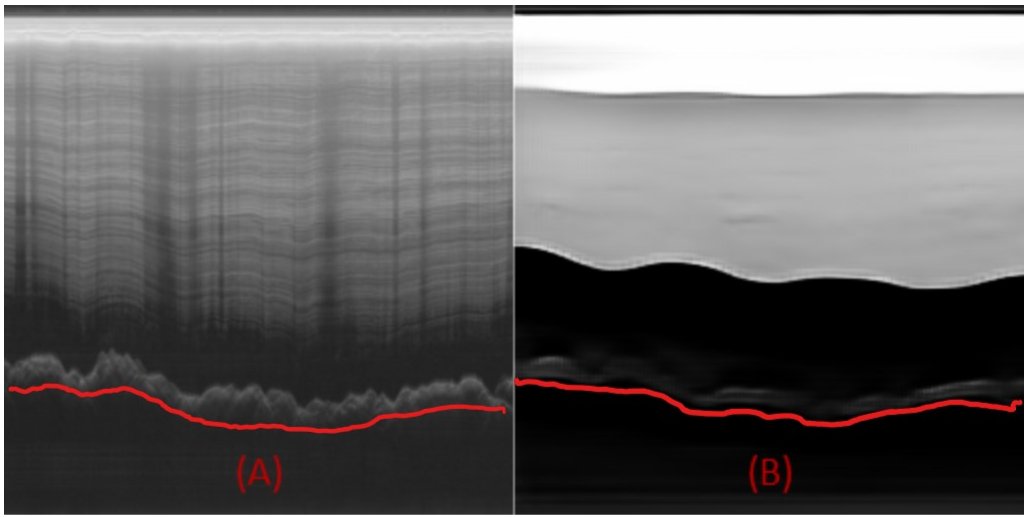

Figure 6: IR1HI1B_2009006_MCM_JKB1a_EDMC01a_000_HIGH. Follows the same format as figure 4.

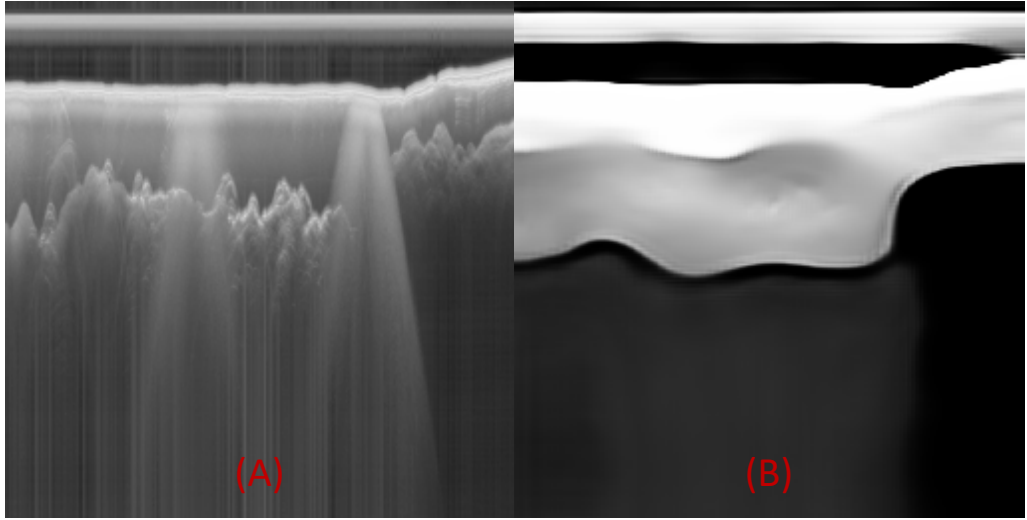

Figure 7: IR1HI1B_2009008_ICP1_JKB1a_F09T01b_005_HIGH. Follows the same format as figure 4.

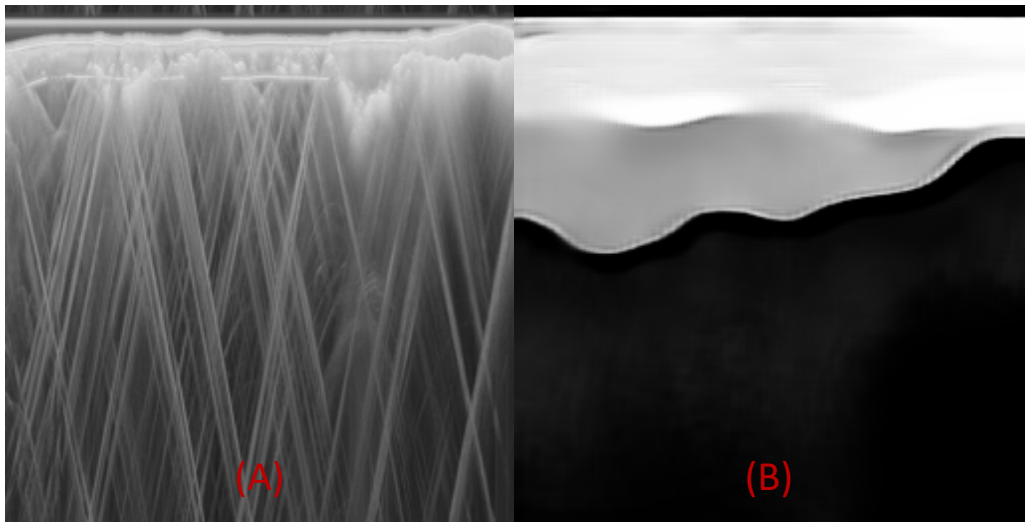

Figure 8: IR1HI1B_2009007_ALG_JKB1a_Y06a_000_HIGH. Follows the same format as figure 4.

### 3.2 Model Accuracy

MSE and PSNR are calculated using images translated from domain X to Y, exemplified by A and B respectively in Figures 5 through 8. We take a lower MSE to signify less distortion in the translation. Since the end goal was an image that was visibly different, an MSE of 0 would be undesirable. However, high values of the MSE would indicate that the model is distorting the image significantly. As the result should be identical in underlying structure to the noisy image, a target MSE was estimated to be above 0.01 and below 0.1 through visual analysis of intermediate results which fell outside these bounds.

The average MSE after translating the real-world data was 0.049 (Table 1), which suggests the model is not removing excess data. The average PSNR for the denoised images was 61.262 dB. Since PSNR is a direct translation of MSE, an MSE range of 0.01 to 0.1 grayscale equates to a PSNR range of 58.1 dB to 68.1. Our resulting average value of 61.26 is thus favorable as it is within our acceptable range.

| Data | Mean-Square-Error (grayscale) | Peak Signal-to-Noise Ratio (dB) |
|---|---|---|
| F09T01b_005 | 0.049 | 61.262 |
| _DGC02a_001_ | 0.059 | 60.402 |
| WLKX01a_005 | 0.046 | 61.480 |
| F09T01b_006 | 0.043 | 61.836 |
| X02a_000 | 0.045 | 61.604 |
| EDMC01a_000 | 0.047 | 61.438 |
| EDMC01a_001 | 0.056 | 60.668 |
| Average | **0.049** | **61.262** |

Table 1: Quantitative scores for Peak Signal-to-Noise Ratio (PSNR) and Mean-Square-Error (MSE) measurements after transform of noisy domain data into clean domain. This represents the amount the input data changes in the standard use case of de-noising.

## 4. Discussion

The model is most effective at translating data with minimal diagonal strip noise such as in Figures 5 and 6. These images are very similar in structure to the synthetic data as they contain relatively smooth layers with little to no diagonal noise obscuring the layer boundaries. We hypothesize that the model handles translating this data especially well due to clearer boundaries between layers.

In contrast, the model is less effective at translating data with significant diagonal noise such as in Figures 7 and 8. This is due to a substantial amount of angled noise that obfuscates the glacial layer boundaries. However, the results from this data type are still usable since the model can track the continuity of layer boundaries. Even though the images have less usable data due to the increased amount of noise, the model can interpolate the missing information and highlight layers. In contrast, filtering techniques eliminate the diagonal noise without any interpolation, leaving the images to human interpretation. Additionally, due to the difficulty of filtering or manually interpreting images of this type, FCNs cannot be accurately trained to parse them either.

A notable flaw is that the model removes detail at the ice-bedrock/ice-water transition and adds a faint layer above it. This happens because of inconsistent transition representation in the synthetic data. In determining glacial layer height, having a well-defined transition is unimportant. However, if such details are relevant in other applications, the issue could be mitigated by generating synthetic data that more closely represents the transitions. Additionally, there is visible loss of detail in the final model, but earlier checkpoints were still viable and could be used when more detail is required.

**5. Conclusion**

In this paper, we illustrate the effectiveness of a CycleGAN to translate RES data from a noisy domain to an automatically interpreted domain. The CycleGAN architecture is an intuitive way to approach this task due to the lack of paired data. We have demonstrated, both by quantitative and qualitative means, that a CycleGAN trained with unpaired synthetic and real-world data can successfully identify layer boundaries and highlight layers in noisy RES records without a large loss of detail or significant distortion.

While the CycleGAN is less effective at translating when there is a large amount of diagonal noise, it still outperforms existing techniques. We believe that significant improvement in this area could be made with the use of a WavCycleGAN (Song et al. 2020). WavCycleGANs utilize a wavelet subband domain learning scheme for translating without sacrificing high frequency components such as edges and detail information, which is something that can enhance results if implemented. Furthermore, we believe that this technique could be applied to three-dimensional analysis in various fields.

We recommend that CycleGAN models trained with synthetic and real-world data be applied in fields where strip noise is an issue, such as in remote sensing data, audio data, and optical data. In the future, there will be an increased reliance on data-driven systems, and filtering techniques, which require significant manual interpretation, will become impractical. Finding ways of automating tasks such as strip noise removal reduces the time required for data cleanup and allows for more analysis. Therefore, we believe that further exploration into CycleGANs for noise removal and further analysis could yield quick, powerful, and effective tools that make data-usage more efficient.